# Arduino Sensor Integrated Drone for Weather Indices: A Prototype for Pre-flight Preparation

**Theodore Karachalios[1], Dimitris Kanellopoulos[2], Fotis Lazarinis[3]**

[1]Hellenic Open University, GR 26335, Patras, Greece. theodoros.karachalios@haf.gr
[2]Department of Mathematics, University of Patras, GR 26500, Greece. d_kan2006@yahoo.gr
[3]Hellenic Open University, GR 26335, Patras, Greece.  fotis.lazarinis@ac.eap.gr

**Abstract**: Commercial weather stations can effectively collect weather data for a specified area. However, their ground sensors limit the amount of data that can be logged, thus failing to collect precise meteorological data in a local area such as a micro-scale region. This happens because weather conditions at a micro-scale region can vary greatly even with small altitude changes. For now, drone operators must check the local weather conditions to ensure a safe and successful flight. This task is often a part of pre-flight preparations. Since flight conditions (and most important flight safety) are greatly affected by weather, drone operators need a more accurate localized weather map reading for the flight area. In this paper, we present the **A**rduino **S**ensor **I**ntegrated **D**rone (ASID) with a built-in meteorological station that logs the weather conditions in the vertical area where the drone will be deployed. ASID is an autonomous drone-based system that monitors weather conditions for pre-flight preparation. The operation of the ASID system is based on the Arduino microcontroller running automatic flight profiles to record meteorological data such as temperature, barometric pressure, humidity, etc. The Arduino microcontroller also takes photos of the horizon for an objective assessment of the visibility, the base, and the number of clouds.

## Introduction

Local meteorological conditions are usually assessed by a brief look at the sky. However, this is insufficient for optimal flight planning. Factors such as wind speed and temperature may vary with the altitude, and thus must be taken into account in all flights. Existing weather models have difficulty in predicting with accuracy local weather phenomena such as fog, high fog, and thunderstorms, due to insufficient coverage of measurement data in the mid and lower levels of the atmosphere [9]. Pre-flight preparation is required for all flights. One stage of pre-flight preparation is *Weather Briefing* that involves determining forecast and actual weather conditions for the route planned and for selected airfields along the route. En-route weather comprises forecast winds and temperatures at cruising levels along the route together with forecasts of en-route weather conditions, especially cloud conditions, and any associated turbulence and/or icing. This information is depicted on special charts. In an airport, one task of pre-flight preparation is to monitor weather conditions because these can affect aviation. Weather conditions/variables in an airport can be prevailed by capturing special data through sensors and processing these data in a computer ground station. Such data are 1) wind direction and intensity; 2) freezing level; 3) dew point; 4) barometric pressure; 5) cloud base and roof; 6) quantity of clouds; 7) discomfort index; 8) visibility; and 9) temperature. The monitoring of weather data can give vivid and exact information for pre-flight purposes and goes beyond weather stations. An Unmanned Aerial Vehicle (UAV a.k.a. drone) equipped with sensors can measure selected meteorological parameters and its current position. In this sense, drones can be exploited during the pre-flight preparation process [17].

Nowadays, the development of drone capabilities has disseminated its influence in many domains, such as defense, search and rescue, agriculture, manufacturing, healthcare services [8], vacant parking space detection [13], and environmental surveillance to execute complex industrial functions [1][7]. Mao et al. [9] presented the Smart Arduino Drone that monitors the weather conditions on a rice field, tea farm, maize, or vegetable fields as these crops require different conditions for optimal productivity. Also, drones have been used to investigate microclimate variation at a small scale and from a pedestrian perspective in urban landscapes [14]. Urban landscapes are strongly characterized by a high level of heterogeneity and complex morphology. This characteristic is responsible for the diversification of microclimate conditions even within the same city and not only by comparing urban conditions and rural surroundings.

In this paper, we present an innovative drone-based system specifically designed for monitoring weather conditions at a microclimate scale. The newly developed system is called *Arduino Sensor Integrated Drone* (ASID) and logs weather variables such as temperature, and humidity and captures images of the horizon in different altitudes. The weather data are synchronized to a ground computer station and are displayed to the user through an application programmed.

### Related Work

As part of pre-flight preparations, drone operators must check the local weather conditions to ensure a safe and successful flight. While commercial weather stations can effectively collect data for a specified area at the macro-scale, weather conditions in that area at the micro-scale can vary greatly. Since flight conditions can be greatly affected by weather conditions, drone operators need a more accurate localized weather map reading for the area of flight. Maurer et al. [10] proposed the *Weather Box* that creates such a localized map in a network of battery-powered sensor modules to provide drone users with the required information via a website and application. This product allows operators to quickly decide whether the conditions are suitable for a safe drone flight.

Many Radiosonde instruments also measure meteorological parameters such as air pressure, temperature, and humidity [3]. The most famous instruments are the VIZ; Space Data Corp.; Chinese GZZ; Japanese RS2-80; Russian RKZ, MARS, and A-22; and Vaisala RS80, RS 12/15, and RS18/21 radiosondes. These instruments have different error and response characteristics. A Radiosonde instrument is often carried by a free-flying balloon up to a height of about 30 km above sea level (asl). Every 30 secs, the measured values are transmitted to the aerological station using a shortwave transmitter. The measurements are carried out usually twice a day. The spatial resolution of the station's distribution is sparse, usually around 250 km or more in Europe [4].

Few studies consider how meteorological parameters can be sensed from a drone. These studies are mostly covering aspects related only to the wind [12][18]. Also, the drones were initially equipped with a single frequency GPS receiver for measuring the location of the drone. For example, the *Meteodrone* measures the meteorological parameters, wind, and position of a drone. The Meteodrone (www.meteomatics.com) is equipped with sensors that measure temperature, wind speed, wind direction, dew point temperature, relative humidity, and air pressure with a sampling rate of 250 ms [19]. The drone size is 70cm x 70cm and its weight is about 1.5 Kg. The maximum climb rate is 10 m/sec, and the theoretical maximum altitude is 3 km above the ground. The drone has a Beyond Visual Line-Of-Sight (BVLOS) permission issued by the Federal Office of Civil Aviation (FOCA), which means that it can fly within clouds and in fog. The Meteodrone has a water-resistant certificate IP64, and thus, it is also capable of flying in the rain. The main limitation is the severe wind: the drone cannot be operated with wind exceeding 50 km/h.

From another perspective, weather conditions are pertinent to drone (UAV) routing because wind speed and wind direction could agitate the travel speed of the UAV, and the air density, appeased by the temperature in the atmosphere, affect battery performance [2]. The majority of the current state of research has paid less attention to weather factors and ignores the impact of weather on performance [5][6][16]. Existing literature has sporadically considered wind conditions on energy consumption and the use of that information in planning the missions [11][15]. The studies have assumed constant wind speed and direction and have used linear approximations for energy consumption, giving less focus on the weather [2].

### ASID: the Arduino Sensor Integrated Drone

As shown in Fig. 1, The *Arduino Sensor Integrated Drone* (ASID) is a QUAD X-shaped quadcopter controlled by an ArduPilot Mega (APM) microcontroller.

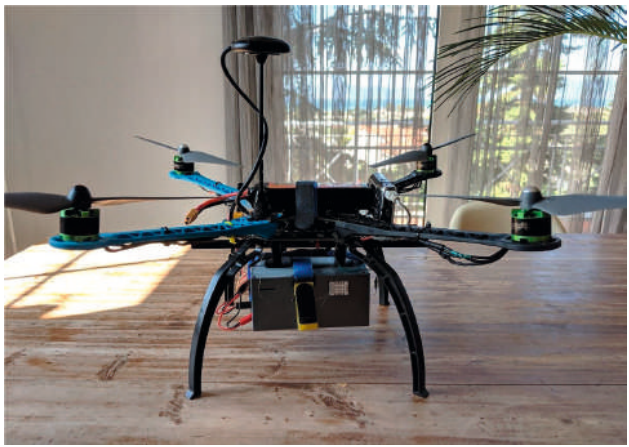

*Figure 1.* The ASID

ASID flies and records weather data. Then, it sends directly this information to the user (pilot) on a computer system located on the ground. For completing this task, ASID has a build-in meteorological station. ASID has been designed to quickly log the key weather parameters to making a successful pre-flight preparation. It is been built as hands-off, easy to use, and as industry-specific as possible. This smart drone is a big boost to pilots because it can completely transform and revolutionize the aviation industry. An important characteristic of ASID (and of every drone) is the *Thrust/Weight* (T/W) ratio because the higher this ratio is the more agile and fast the drone is. The T/W ratio is calculated as follows:

$$\frac{T}{W} = \frac{Maximum\ Thrust \times Number\ of\ Motors}{Total\ Weight} \quad (1)$$

If the *Total Weight* increases, the *T/W* ratio decreases, and thus the drone performance is decreased. During the drone flight, the T/W ratio changes depending on the current altitude and temperature. The thrust depends mainly on the density of the air and decreases proportionally as we climb higher. The maximum T/W ratio appears at sea level and with the motors running at maximum intensity. The T/W ratio is decreasing as the drone climbs because the air density decreases. On a usual day (Temperature=15C) at 20,000 feet, the air density is 0.660 Kg/m$^3$, while at sea level, it is 1,225 Kg/m$^3$. If we assume that a drone is a quadcopter and weighs 2 Kgr, the thrust at 20,000 feet should be 3 Kgr. Depending on the day temperature, we can calculate that the required static thrust is 5,586 gr. Therefore, we have 5600/4 (1400) grams distributed in each engine when the drone is a quadcopter. Finally, if we increase the battery capacity (and thus the total weight), the T/W ratio is reduced.

### Flight Characteristics

ASID has the following flight characteristics:
- *The maximum flight altitude for ASID is 20,000 feet.* This altitude is close enough to the tropopause (i.e., the area where the weather stops). Thus, ASID does not need to fly higher to log data. At the maximum flight altitude, the T/W ratio becomes 1 and the drone reaches its design limit. Apart from the T/W ratio, the *endurance* of a drone in the maximum reachable altitude is important as there is a possibility that the battery capacity will not be sufficient for the time required to climb there.
- *The wind limit for ASID is 9Bft (or 75km/h).* ASID can fly in winds up to 9 Beaufort. This wind limit is calculated on the ground that with winds above 9 Beaufort, no flights are operated. For a drone to fly with the wind limit, its progressive speed must be greater than or equal to the wind speed. To calculate the progressive speed, we apply the following Equation (2). The calculation is complicated since we have to calculate the thrust loss that exists from the inclination of the fuselage to be able to move.

$$F = 1.225 \frac{\pi(0.0254 \cdot d)^2}{4} \left[ \left( RPM_{prop} \cdot 0.0254 \cdot pitch \cdot \frac{1min}{60sec} \right)^2 - \left( RPM_{prop} \cdot 0.0254 \cdot pitch \cdot \frac{1min}{60sec} \right) v_o \right] \left( \frac{d}{3.29546 \cdot pitch} \right)^{1.5}$$

(2)

- *ASID can take off with greater wind intensity.* However, its landing point will be shifted in the direction of the wind. For example, for a 3-minute flight with winds of 12 Beaufort, ASID will land 2 km away from its take-off point.
- *The maximum climb rate near the ground can reach 120 ft/sec* and decreases as the drone increases its altitude. Near the maximum altitude, the calculated climb rate would be almost 55 ft/s. With these climb rates, the drone will need almost 7-8 minutes to reach its maximum altitude. This time is needed to calculate the battery capacity. Theoretically, the maximum ascent speed in ft/s is calculated if we subtract the weight as well as the air resistance in the frame from the maximum thrust. This speed changes during the ascent as the thrust decreases, the weight, and the air resistance decrease. The reduction of the last two is much smaller than the impulse, so the speed decreases during the ascent.
- *ASID is waterproof.* We selected appropriate materials to seal the drone in the rain. DC Brushless Motor motors are waterproof and work even in water. Regarding Electronic Speed Controller (ESC), we selected materials that meet the IP-68 standard and thus can withstand the maximum volume of water that we statistically observe in Greece. The meteorological station is installed in an electrical box that has IP-68 sealing.
- *ASID supports stability* that depends on the correct weight distribution. We selected specific motors so that with half thrust output motors can hold ASID in the air in the HOVER position.
- *ASID supports reliability*: In terms of reliability, the most critical material in construction is the engines. By choosing quality materials we can achieve more than 160 hours of Mean Time Between Failures (MTBF) which translates to about 960 flights without expected failure.
- *A user can create a flight profile that can run automatically.*
- *The meteorological station weighs 200 gr and is installed inside ASID.*

**Technical Specifications**

ASID has a flight controller, a GPS receiver, speed controllers, a telemetry system, propellers, and a Li-PO battery. The arrangement of 4 motors in our QUAD X—shaped quadcopter is shown in Fig. 2.

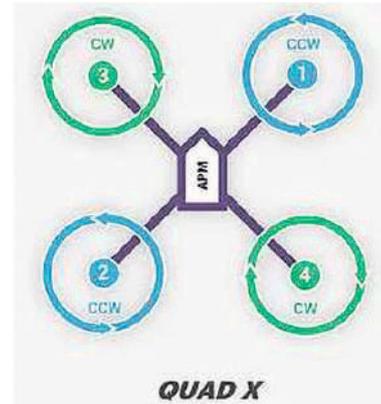

*Figure 2.* Arrangement of 4 motors in Quad X

*The Lithium-polymer (Li-PO) battery* provides greater energy capacity for the same size than other types of lithium batteries and can be used in applications where weight is a critical feature. The main characteristics of a Li-PO battery are its operating voltage, capacitance, and discharge rate of the voltage. The LiPo battery has a nominal voltage of 3.7 Volts. Increasing the voltage, the efficiency of the drone engines can be improved. A LiPo battery used in drones often has a capacity of 5000 mAh as it provides a good capacity/weight ratio. Such a battery delivers five Amps in 1 hour. The charging rate shows how fast the battery can deliver its energy without being damaged. The charging rate indicator along with the battery capacity allows us to calculate the *maximum load* (*Charging rate X capacity*) we can request without overheating it and thus damaged. For example, a battery with a capacity of 5000 mAh and a charging rate of 50C can give us a load of 250 Amp without overheating.

*Electric motors*: ASID uses electric motors as they have high torque and small weight. The torque of an electric motor is necessary for the instant change in the engine speed for the drone to be agile. High torque also simplifies the manufacturing of the engine since it does not need transmission gears, which means less weight, and thus a higher thrust-to-weight ratio. The main characteristics of electric motors are the diameter, the speed per Volt (KV), and the operating voltage. Electric motors are characterized by a four-digit number XXYY that represents their dimensions. The first 2 digits represent their diameter while the last 2 digits represent their height. For example, a 2204 electric motor has a diameter of 22 mm and a height of 04 mm. Speed per Volt (KV) of a motor indicates the number of rounds/per minute with a voltage of 1Volt. This measurement is load-free which means that the actual speed depends on the size of the propeller. KV determines the maximum speed that an engine can achieve. Low KV means that the motor has more torque so more energy per rotation, a motor with high KV will rotate faster but will have less torque. Each manufacturer provides us with the capabilities of each motor depending on the operating voltage as well as the size of the propeller. The flight of the ASID drone is achieved without any kind of rudder since it is based on the asymmetric thrust it creates with the four engines.

*The 3DR Radio Telemetry 915 MHz component:* Using the ground station (mobile phone or laptop), the flight operator monitors and records all flight parameters by using the radio telemetry component. Using the telemetry component, the pilot checks the drone and sends a new flight plan to be performed. The telemetry system consists of two RF transceivers at 915 MHz or 433 MHz and has a bandwidth of 250Kbs up to 300 Mbps. The 3DR Radio Telemetry 915 MHz component (shown in Fig. 3) contains two transceivers, where one is with a USB plug for connection to a computer or mobile phone via OTG.

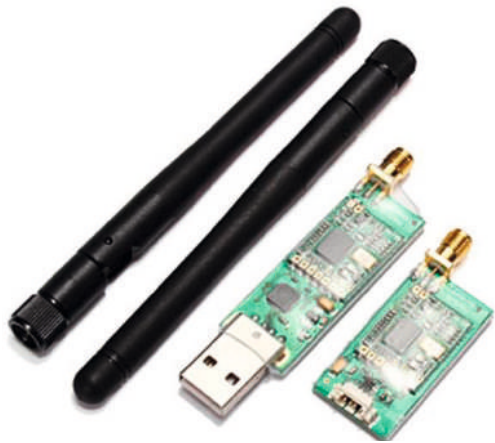

*Figure 3.* The 3DR Radio Telemetry 915 MHz component

The *Flight Controller* (FC) of ASID (shown in Fig. 4) is an Arduino microcontroller that performs all the calculations required for the flights. Sensors are connected to the FC receiver and transmit their sensed weather data. The FC sends the corresponding signals to the Electronic Speed Controller (ESC) that controls the speed of the motors and by using the asymmetric thrust controls it. For the proper FC operation, a constant power supply of 5V is required. Due to the different operating voltage of the Li-Po battery, a special Battery Elimination Circuit (BEC) circuit is used that delivers a constant voltage of 5V DC to the FC.

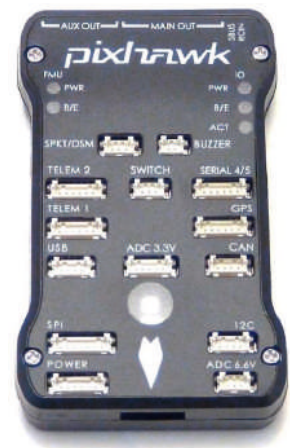

*Figure 4.* Flight Controller PixHawk

The *Electronic Speed Controller* (ESC) regulates the speed of an electric motor based on the signals it receives from the FC. ESCs are categorized based on the maximum current Ampere they can provide and the software used. The ESCs are connected to the LiPO battery and to the FC that gives the corresponding signal.

*The Arduino central processor board and the ATmega2560 microcontroller*: The Arduino central processor board has low power consumption and minimal manufacturing cost. It includes an 8-bit microcontroller (ATmega2560) that has 54 digital inputs and 16 analog inputs. The board has also some extra cards (known as "shields") which offer more features to the microcontroller. These cards can provide additional connectivity to the microcontroller, such as Ethernet network (Ethernet shield), wireless WIFI network, wireless serial communication with Bluetooth Technology (BT Shield), etc. In our project, we used the ARDUINO MEGA 2560 REV3 (shown in Fig. 5). The ATmega2560 microcontroller is an open-source microprocessor. This means that it can be programmed in Wiring (a variant of the C++ programming language). For the implementation of ASID, we used the functions and sensors of the ArduPilot board to create an automatic navigation system.

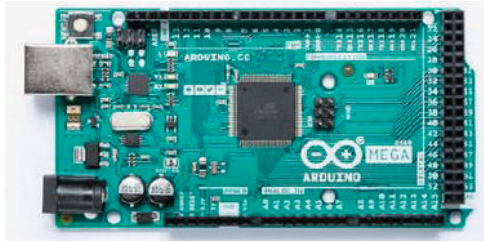

*Figure 5. The ATmega2560 Microcontroller*

*Arduino Software:* The Arduino has some portals that function as inputs or outputs. The functionality of these portals is managed by writing code in the Arduino IDE programming environment (Arduino 1.8.5 ver. for Windows 10).

### The Built-in Weather Station

Figure 6 shows how the weather station operates.
1. Initially, ASID logs temperature, humidity, and air pressure values on the ground multiple times to accurately calculate the median values of these variables.
2. Then, ASID will take-off to log the same weather variables in the air at a specific height interval.
3. As soon as the ASID system lands, an HTTP fileserver is created to synchronize the logs with the presentation program running on the ground station (laptop).

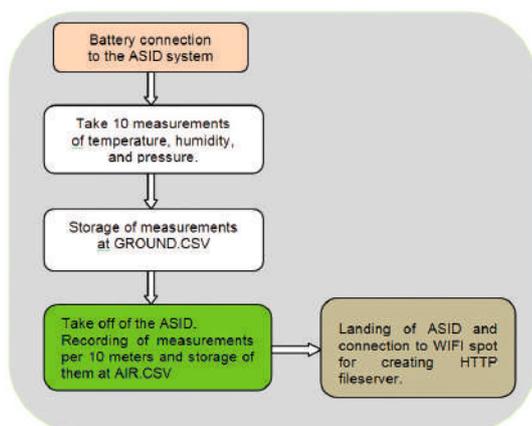

*Figure 6. The operation diagram of ASID*

All the captured data are transmitted to the ground station via a wireless link (2.4 GHz). The display of data on the laptop is obtained using a Java program. Moreover, a camera receives commands from the Flight Controller and takes photos whenever we want. Photos from the camera can either be transferred with a card reader to the laptop or with the use of an Sd air-fi card that through a WIFI connection the data can be synchronized.

The Arduino source code for the operation of the meteorological station was written in Arduino IDE 1.85.

**Setup Method:** The setup method performs the initialization of all sensors and calculates the Mean Sea Level Pressure that is stored in the variable named SLpressure_HPA. The value of this variable is crucial for the calculation of the rest of the logged variables.

```
1.  void setup() {
2.  Serial.begin(115200);
3.  pinMode(buzz, OUTPUT);
4.  SD.begin(chipSelect);
5.  delay(2000);
6.  bmp280.begin(0x76);
7.  rtc.begin();
8.  dht.begin();
9.  SLpressure_mB = GetMSLP();
10. SLpressure_HPA = SLpressure_mB / 100;
11. }
```

### Creating a Flight Profile

We planned a certain flight and created a flight profile using the Mission Planner application (Fig. 7). Mission Planner application transmits (uploads) the data of this flight profile to the FC to execute this profile automatically. The flight profile has the following commands:
1. Take-off and ascent to 10 meters AGL (a height above ground level)
2. Rotate to 90 degrees (East)
3. Capture Image
4. Rotate 180 degrees (South)
5. Capture Image
6. Rotate to 270 degrees (West)
7. Capture Image
8. Rotate 360 degrees (North)
9. Capture Image
10. Rise to 20 meters AGL

These tasks are repeated until the drone climbs to the desired altitude. Then, the drone will continue descending up to 10 meters and after a few seconds, it will land automatically.

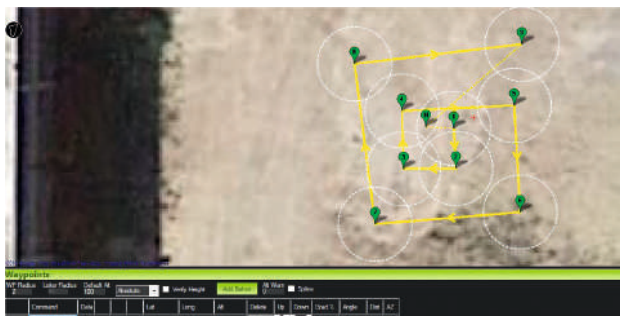

*Figure 7.* (a) The Mission Planner (points).

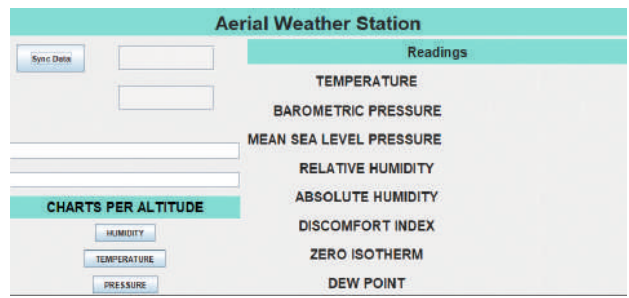

*Figure 8.* The Aerial Weather Station application

clicks the SYNC button, the files from the Fileserver are synchronized. Then, the necessary data are processed and displayed in the main window. The time of data collection and data process is also presented.

## Conclusion

The autonomous weather record system provides better weather forecasting and monitoring services in the aviation industry. Shortly, we aim to add new important features in the ASID system such as standalone charging, 360 cameras, lightning sensor, infrared (IR) imaging for fire detection, weather reports on websites the automatic upload of Meteorological Terminal Air Reports (METAR), etc.

*Figure 7.* (b) Mission Planner (commands)

In Figure 7a, the points are depicted not to be vertical as for the presentation of the drone movements to be more understandable to the reader. The software also allows us to activate the Shutter when we need to capture an image. This signal is transmitted via a special cable between the camera and the FC. Due to the speed of the camera, a delay of 10 ms has been inserted between the changes of course.

### The "Aerial Weather Station" Application

We developed an application called "Aerial Weather Station" (shown in Fig. 8). We implemented the Graphical User Interface (GUI) of this application by using Java language with IDE intellijIDEA. The GUI of our application displays the following graphs: (1) Height/temperature graph; (2) Height/humidity graph; (3) Height/pressure graph; (4) Freezing Level; (5) Dew Point; (6) Discomfort Index on the surface; (7) Barometric pressure on the surface; (8) Freezing Level. The main window of the application also includes a button for synchronizing logs as well as for exporting calculations. When the user

**Abbreviations**

| | |
|---|---|
| AGL: | A height Above Ground Level |
| ASID: | Arduino Sensor Integrated Drone |
| ESC: | Electronic Speed Control |
| FC: | Flight Controller |
| FL: | Flight Level |
| GPS: | Global Positioning System |
| GUI: | Graphical User Interface |
| LiPo: | Lithium Polymer |
| METAR: | Meteorological Terminal Air Report |
| MTBF: | Mean time between failures |
| T/W: | Thrust/Weight ratio |
| UAV: | Unmanned Aerial Vehicle (a.k.a. drone) |

**APPENDIX A CAN BE USED ONLY FOR THE REVIEW PROCESS.**

## APPENDIX A: The source code

```
1.  //ΕΙΣΑΓΩΓΗ LIBRARIES//
2.  #include "DHT.h"
3.  #include < DS3231.h >
4.  #include < SD.h >
5.  #include < stdlib.h >
6.  #include < SPI.h >
7.  #include < SD.h >
8.  #include < Wire.h >
9.  #include "WiFiEsp.h"
10. #include < Adafruit_Sensor.h >
11. #include < Adafruit_BMP280.h >
12. //ΟΡΙΣΜΟΣ PIN//
13. #define DHTPIN 8
14. #define DHTTYPE DHT22
15. Adafruit_BMP280 bmp280;
16. DHT dht(DHTPIN, DHTTYPE);
17. DS3231 rtc(SDA, SCL);
18. WiFiEspServer server(80);
19. const int buzz = 7;
```

```
20. //ΟΡΙΣΜΟΣ ΜΕΤΑΒΛΗΤΩΝ//
21. String readString;
22. const int chipSelect = 53; //PIN SDCARD
23. float temperature_DHT;        // Θερμοκρασία (C)
24. float heat;                   // Δείκτης δυσφορίας
25. float humidity_DHT;           // Υγρασία σε %
26. float altimeter;              //Υψόμετρο
27. float CalAltimeter;           // Διορθωμένο υψόμετρο με βάση την MSLP
28. float SLpressure_HPA;         // MSLP σε hPa
29. float pressure;               // Βαρομετρική πίεση σε Pa
30. float pressure_HPA;           // Βαρομετρική πίεση σε hPa
31. float PressureCorrection = 0.995; // Μεταβλητή διόρθωσης βαρομετρικής πίεσης
32. float PressureCorrected;      // Διορθωμένη βαρομετρική πίεση
33. float SLpressure_mB;          // Βαρομετρική Επιφανείας
34. int Interval = 5;             // Υψομετρικό διάστημα καταγραφής δεδομένων στο air.csv
35. int run = 0;                  // Μεταβλητή για το αν έχει γίνει log στο ground.csv
36. int listenint = 0;            // Μεταβλητή για το αν είναι έτοιμος ο fileserver
37. int ELEVATION = 45;           //Υψόμετρο που βρισκόμαστε για υπολογισμό της MSLP
38. char ssid[] = "diktio";       // όνομα SSID wifi hotspot
39. char pass[] = "2610333918";   // κωδικός wifi
40. int status = WL_IDLE_STATUS;  // Κατάσταση σύνδεσης Esp8266
41. //0.Μέθοδος δημιουργίας ήχου
42. void buzzer(int time) {
43.     tone(buzz, 1000);         // Send 1KHz sound signal...
44.     delay(time);              // ...for 1 sec
45.     noTone(buzz);             // Stop sound...
46. }
47. //1.Μέθοδος Δημιουργίας Http Fileserver//
48. void MakeServer() {
49.     Serial1.begin(115200);
50.     WiFi.init( & Serial1);
51.     while (status != WL_CONNECTED) {
52.         Serial.print("Προσπάθεια για σύνδεση στο WPA SSID: ");
53.         Serial.println(ssid);
54.         status = WiFi.begin(ssid, pass);
55. }
56.     Serial.println("Επιτυχής Σύνδεση");
57.     printWifiStatus();
58.     server.begin();
59.     listenint = 1;
60. }
61. //2.Μέθοδος HTTP send του ground.csv//
62. void DownloadGround() {
63.     WiFiEspClient client = server.available();
64.     client.println("HTTP/1.1 200 OK");
65.     client.println("Content-Type: text/csv");
66.     client.println("Content-Disposition: attachment; filename=\"ground.csv\"");
67.     client.println("Connection: close");
68.     client.println();
69.     File myFile = SD.open("ground.csv");
70.     if (myFile) {
71.         Serial.println("file opened"); //send new page
72.         byte clientBuf[1760];
73.         int clientCount = 0;
74.     while (myFile.available()) {
75.         clientBuf[clientCount] = myFile.read();
76.         clientCount++;
77.         if (clientCount > 1759) {
78.             client.write(clientBuf, 1760);
79.         clientCount = 0;
80.         }
81.     }
82.     if (clientCount > 0) client.write(clientBuf, clientCount);
83.     myFile.close();
84.     client.stop();
85.     readString = "";
86.     Serial.println("Removing air.csv...");
87.     SD.remove("air.csv");
88.     Serial.println("Removing ground.csv...");
89.     SD.remove("ground.csv");
90. }
```

```
91.     }
92.  //3.Μέθοδος HTTP send του air.csv//
93.  void DownloadAir() {
94.     WiFiEspClient client = server.available();
95.     client.println("HTTP/1.1 200 OK");
96.     client.println("Content-Type: text/csv");
97.     client.println("Content-Disposition: attachment; filename=\"air.csv\"");
98.     client.println("Connection: close");
99.     client.println();
100.           File myFile = SD.open("air.csv");
101.           if (myFile) {
102.                   Serial.println("file opened");
103.                   byte clientBuf[1760];
104.                   int clientCount = 0;
105.                   while (myFile.available()) {
106.                           clientBuf[clientCount] = myFile.read();
107.                           clientCount++;
108.                           if (clientCount > 1759) {
109.                                   client.write(clientBuf, 1760);
110.                                   clientCount = 0;
111.                           }
112.                   }
113.           if (clientCount > 0) client.write(clientBuf, clientCount);
114.           myFile.close();
115.           client.stop();
116.           readString = "";
117.           delay(5000);
118.           client.stop();
119.           return;
120.           }
121.  }
122.  //4.Μέθοδος αναμονής του HTTP fileserver για πελάτη//
123.  void listen() {
124.     WiFiEspClient client = server.available();
125.           if (client) {
126.                   Serial.println("New client");
127.                   boolean currentLineIsBlank = true;
128.                   while (client.connected()) {
129.                           if (client.available()) {
130.                                   char c = client.read();
131.                                   readString += c;
132.                                   if (c == '\n') {
133.                                     Serial.println(readString.indexOf("air"));
134.                                     Serial.println(readString.indexOf("ground"));
135.                                     if (readString.indexOf("air") > 5)  DownloadAir();
136.                                     else DownloadGround();
137.                                   }
138.                           }
139.                   }
140.     }
141.  }
142.  //5.Μέθοδος εκτύπωσης στοιχείων σύνδεσης//
143.  void printWifiStatus() {
144.     Serial.print("SSID: ");
145.     Serial.println(WiFi.SSID());
146.     IPAddress ip = WiFi.localIP();
147.     Serial.print("IP Address: ");
148.     Serial.println(ip);
149.     Serial.println();
150.     Serial.print("To see this page in action, open a browser to http://");
151.     Serial.println(ip);
152.     Serial.println();
153.  }
154.  //6.Μέθοδος υπολογισμού της MSLP//
155.  float GetMSLP() {
156.     pressure = bmp280.readPressure();
157.     PressureCorrected = pressure * PressureCorrection;
158.     float SLpressure_mB = (((PressureCorrected * 100.0) / pow((1 - ((float)(ELEVATION)) / 44330), 5.255)) / 100.0);
159.     return SLpressure_mB;
160.  }
```

```
161. //7.Μέθοδος καταγραφής δεδομένων στο air.csv//
162. void LogAir() {
163.   humidity_DHT = dht.readHumidity();
164.   temperature_DHT = dht.readTemperature();
165.   heat = dht.computeHeatIndex(temperature_DHT, humidity_DHT, false);
166.   pressure = bmp280.readPressure();
167.   PressureCorrected = pressure * PressureCorrection;
168.   pressure_HPA = PressureCorrected / 100;
169.   SLpressure_HPA = SLpressure_mB / 100;
170.   altimeter = bmp280.readAltitude(SLpressure_mB);
171.   CalAltimeter = (SLpressure_HPA - pressure_HPA) / 0.12;
172.   File dataFile = SD.open("air.csv", FILE_WRITE);
173.   if (dataFile) {
174.           dataFile.print(rtc.getDateStr()); //Log Date
175.           dataFile.print(","); //Next Column
176.           dataFile.print(rtc.getTimeStr()); //Log Time
177.           dataFile.print(","); //Next collumn
178.           dataFile.print(temperature_DHT, 1);
179.           dataFile.print(",");
180.           dataFile.print(humidity_DHT, 1);
181.           dataFile.print(",");
182.           dataFile.print(heat, 1);
183.           dataFile.print(",");
184.           dataFile.print(pressure_HPA, 2);
185.           dataFile.print(",");
186.           dataFile.print(CalAltimeter, 2);
187.           dataFile.print(",");
188.           dataFile.println();
189.           dataFile.close();
190.   } else Serial.println("Αποτυχία εγγραφής");
191. }
192. //8.Μέθοδος καταγραφής δεδομένων στο ground.csv//
193. void LogGround() {
194.   humidity_DHT = dht.readHumidity();
195.   temperature_DHT = dht.readTemperature();
196.   heat = dht.computeHeatIndex(temperature_DHT, humidity_DHT, false);
197.   pressure = bmp280.readPressure();
198.   PressureCorrected = pressure * PressureCorrection;
199.   pressure_HPA = PressureCorrected / 100;
200.   SLpressure_HPA = SLpressure_mB / 100;
201.   altimeter = bmp280.readAltitude(SLpressure_mB);
202.   CalAltimeter = (SLpressure_HPA - pressure_HPA) / 0.12;
203.   File dataFile = SD.open("ground.csv", FILE_WRITE);
204.   if (dataFile) {
205.           dataFile.print(rtc.getDateStr()); //Log Date
206.           dataFile.print(","); //Next Column
207.           dataFile.print(rtc.getTimeStr()); //Log Time
208.           dataFile.print(","); //Next collumn
209.           dataFile.print(temperature_DHT, 1);
210.           dataFile.print(",");
211.           dataFile.print(humidity_DHT, 1);
212.           dataFile.print(",");
213.           dataFile.print(heat, 1);
214.           dataFile.print(",");
215.           dataFile.print(pressure_HPA, 2);
216.           dataFile.print(",");
217.           dataFile.print(CalAltimeter, 2);
218.           dataFile.print(",");
219.           dataFile.println();
220.           dataFile.close();
221.   } else Serial.println("Αποτυχία εγγραφής");
222. }
223. //9.Αρχικοποίηση των Sensor//
224. void setup() {
225.   Serial.begin(115200);
226.   pinMode(buzz, OUTPUT);
227.   SD.begin(chipSelect);
228.   delay(2000);
229.   bmp280.begin(0x76);
230.   rtc.begin();
231.   dht.begin();
```

```
232.    SLpressure_mB = GetMSLP();
233.    SLpressure_HPA = SLpressure_mB / 100;
234. }
235. //10.Main Loop//
236. void loop() {
237.    if (run == 0) {
238.            for (int i = 0; i <= 5; i++) {
239.                    Serial.println("Log Ground");
240.                    buzzer(500);
241.                    LogGround();
242.                    delay(3000);
243.                    if (i == 5) {
244.                            run = 1;
245.                    }
246.            }
247.    } else {
248.            if (CalAltimeter > Interval) {
249.                    Serial.println("Log Air");
250.                    Serial.println(Interval); //buzzer(2000);
251.                    LogAir();
252.                    Interval = Interval + 5;
253.                    delay(3000);
254.            }
255.    }
256. if (run == 1 && Interval > 35 && listenint == 0) {
257.   MakeServer();
258.   buzzer(5000);
259.   listenint = 1;
260. }
261. if (run == 1 && Interval > 35 && listenint == 1) {
262.         listen();
263.   }
264. }
```

## About the authors

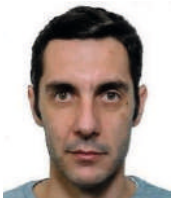

**Theodore Karachalios** holds a BSc in Aviation Science from Hellenic AF Academy. He has also received a MSc in Computer Science from Hellenic Open University in 2018. He is currently working as an Instructor Pilot for H.A.F with more than 2000 HRS in multi-engine fighters. Moreover, he has actively been involved in Flight Safety related projects and Risk Management as he has been a graduate of the International Flight Safety Officer school in NM/USA. Apart from a great experience in manned flights, he is a UAV instructor and currently conducting research and construction of UAV prototypes.

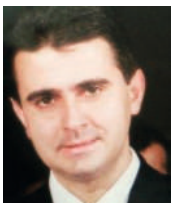

**Dimitris Kanellopoulos** is a member of the Educational Software Development Laboratory in the Department of Mathematics at the University of Patras, Greece. He received a Diploma in Electrical Engineering and a Ph.D. in electrical and computer engineering from the University of Patras. He is a member of the IEEE Technical Committee on multimedia communications. His research interests are multimedia networks and wireless ad hoc networks. He has published numerous articles in refereed journals and conference proceedings, and has edited two books on "Multimedia Networking".

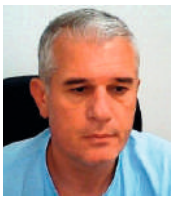

**Fotis Lazarinis** holds a Ph.D. in educational technology from the Teesside University, UK. He works as an adjunct lecturer at the Hellenic Open University. He participates in various research and development projects with a focus on educational technology. He has co-authored more than 50 papers in international journals and conferences.